\documentclass{l4dc2026}
\usepackage{times}
\usepackage{multirow} 

\title[Dynafit]{Minimum distance classification for nonlinear dynamical systems}

\author{
 \Name{Dominique Martinez} \Email{dominique.martinez@univ-amu.fr}\\
 \addr Aix-Marseille University, CNRS, ISM, 13009 Marseille, France
 }

\begin{document}

\maketitle

\begin{abstract}%
We address the problem of classifying trajectories or sequences generated by nonlinear dynamical systems, where each class corresponds to a distinct dynamical system. We propose Dynafit, a kernel-based method that learns a distance metric between training data and the underlying dynamics. New observations are assigned to the class whose dynamics best fit the observations according to the learned metric.
The learning algorithm approximates the Koopman operator, which globally linearizes the dynamics in a (potentially infinite-dimensional) feature space associated with a kernel function. The distance metric is computed in the feature space independently of its dimensionality by exploiting the kernel trick commonly used in machine learning. The kernel function can be tailored to incorporate prior knowledge of the dynamics when available. 
We consider a classical test example, the logistic map as a discrete dynamical system, and derive analytically the kernel function from the polynomial Koopman basis that exactly linearizes the dynamics.
Dynafit is applicable to a wide range of classification tasks involving nonlinear dynamical systems and sensors. We illustrate its effectiveness through three examples: chaos detection in the logistic map, recognition of handwritten dynamical patterns, and classification of visual dynamic textures.
\end{abstract}


\section{Introduction}

Nowadays, the multiplication of sensors and portable devices promote the development of lightweight machine learning (ML) or tiny ML capable of performing sensor data classification at the edge, thereby avoiding the need to transfer data to the cloud \citep{Ravindran_2025, Ray_2022}. Deep learning \citep{LeCun_etal_2015} is not the ideal candidate for classification on embedded systems as it requires a large number of training data combined with lots of computational and memory resources. 
In contrast, traditional ML is less expensive in terms of memory and computation. This is particularly the case for distance-based classifiers, such as k-nearest neighbors \citep{Cover_Hart_1967}, k-means \citep{Lloyd_1982} or Parzen windows \citep{Parzen_1962}, which are among 
the simplest classification algorithms. 
The classification
rule is based on the (dis)similarity, as measured by
a distance function, between test data and stored prototypes from the training set.
Distance-based classifiers have been developed, for static data, using the Euclidean or Mahalanobis distance, and, 
for time series, using the DTW (Dynamic Time Wraping) distance \citep{Abanda_etal_2019}. Kernel machines, such as SVMs (Support Vector Machines) \citep{Burges_1998}, can also be seen as distance-based classifiers as the kernel function can be interpreted as a measure of similarity between data points in transformed feature space \citep{Scholkopf_2000}.  \\

Distance-based classifiers evaluate the similarity between pairs of samples (observed data vs stored prototypes) either in the original or feature space. 
In most cases, the underlying dynamical system is not explicitly taken into account when dealing with sequential data. However, we argue that the generative process behind the data can carry meaningful information for classification. A representative example is pen dynamics in signature verification: unlike the static image of a signature, the motion data is much harder for a skilled forger to reproduce. Therefore, when data originates from a dynamical system, classification could leverage a similarity measure not between individual samples, but between the observed data and the dynamics of the underlying system itself (Figure 1). \\

Sequential data are frequently modeled using state-space representations, for which various distance metrics have been proposed. The Martin distance, for instance, compares two state-space models based on the principal angles between their observability subspaces \citep{Martin_2000,DeCock_DeMoor_2000}. 
Alternatively, the parity space method enables the comparison of a state-space model with observed data by projecting the data onto the parity space — the subspace orthogonal to the observability subspace \citep{Patton_Chen_1991}. However, both approaches rely on prior knowledge of the system dynamics, specifically the state-space matrices. In the absence of an explicit model, data-driven methods have been developed to learn a suitable metric, typically inspired by parity space-based distances \citep{Battistelli_Tesi_2021,Martinez_Boutayeb_2023,Martinez_Boutayeb_2024}. These methods, however, are currently limited to linear systems, and extending them to nonlinear dynamics remains an open challenge.\\

Nonlinear dynamics have traditionally been handled using Jacobian and first-order Taylor series approximation. However, such linearization techniques are inherently local and only valid in the vicinity of fixed points. In contrast, the Koopman operator theory \citep{Koopman_1931,Koopman_Neumann_1932} offers a global linear representation of nonlinear dynamics, albeit in an infinite-dimensional space. Koopman spectral analysis has recently gained renewed interest for the identification and control of nonlinear systems \citep{Mezic_2005, Mezic_2013, Brunton_etal_2016}. For comparing dynamical systems, Mezić and collaborators provide a theoretical framework based on the spectral properties of the Koopman operator \citep{Mezic_Banaszuk_2004, Mezic_2016}. In practice, data-driven methods such as the extended Dynamic Mode Decomposition \citep{Williams_etal_2015a} can approximate the Koopman operator as a set of eigenvalues, eigenfunctions, and modes that span a Koopman-invariant subspace \citep{Brunton_etal_2016}. However, the dimension of this subspace can become intractably large in practice.\\

Kernel methods, by expressing distances through inner products, circumvent the need for explicit mappings into high-dimensional feature spaces \citep{Scholkopf_2000}. 
In this work, we propose Dynafit, a data-driven algorithm for classifying nonlinear dynamical trajectories that leverages kernel methods and Koopman operator approximations. Dynafit classifies trajectories by identifying the minimum distance to the underlying dynamics. It differs from \citep{Wolf_Shashua_2003} and \citep{Vishwanathan_etal_2007}, which compare subspace representations of dynamical systems and emphasize geometric similarity, and from \citep{Williams_etal_2015b}, which analyzes the spectral properties of dynamics in a reproducing kernel Hilbert space.

\begin{figure}[htbp]
\centerline{
\includegraphics[width=12cm]{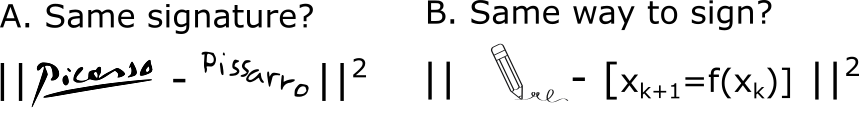}
}
\caption{Illustration of distance-based classification.  A). In traditional classifiers, the distance is a measure of (dis)similarity between a test sample (here the image of Picasso's signature) and a stored prototype (Pissarro's signature). B). In Dynafit, the distance is between a trajectory sample (here the handwritten dynamics) and the dynamical system that has potentially generated the data.  
}
\end{figure}

\section{Problem statement}

Let us consider a discrete-time system represented as 

\begin{eqnarray}
x_{k+1} = f( x_k )
\label{Eq:system1}
\end{eqnarray}
 
with $x \in \mathbb{R} ^{n}$ the $n$-dimensional state and $f$ a nonlinear function. 
A trajectory vector $X \in \mathbb{R} ^{nN}$ consists of a sequence of $N$ states 

\begin{eqnarray}
X = \begin{pmatrix}x_0\\ x_1\\ \cdots\\x_{N-1}\end{pmatrix} %\nonumber
\label{Eq:trajectory_vector}
\end{eqnarray}

The Koopman operator theory \citep{Koopman_1931,Koopman_Neumann_1932,Brunton_etal_2016} provides a linear but infinite-dimensional representation of nonlinear dynamics so that (\ref{Eq:system1}) can be rewritten as  

\begin{eqnarray}
\phi(x_{k+1}) = A \; \phi(x_k)  
\label{Eq:system2}
\end{eqnarray}

with $m \times m$ transition matrix $A$.  The augmented state   
$\phi(x)\in \mathbb{R} ^{m}$ is obtained through a nonlinear transformation $\phi(\cdot)$ onto a high-dimensional ($m>>n$), potentially infinite ($m=\infty$), feature space. 
Eq. (\ref{Eq:system2}) does not represent a linearized version of the dynamics  in the neighborhood of a  fixed point that could be obtained when $A$ is a Jacobian for example. Instead,  it provides a global linear representation of the dynamics in a high-dimensional Hilbert space.
In this feature space, observability matrix $O_b$ maps initial state $\phi(x_0)$ into a $l$-dimensional trajectory vector $\phi(X) \in \mathbb{R} ^{l}$ with $l=mN$ : 

\begin{align}
\phi(X) = O_b \; \phi(x_0) 
\label{Eq:observability_Eq}
\end{align}

\begin{eqnarray}
\hbox{with } \phi(X) = \begin{pmatrix}\phi(x_0)\\ \phi(x_1)\\ \cdots\\\phi(x_{N-1})\end{pmatrix} 
\; \hbox{and } \;  O_b = \begin{pmatrix}I_m\\A\\ A^2\\ \cdots\\A^{N-1}\end{pmatrix} 
\label{Eq:observation_vector}
\end{eqnarray}

where $I_m$ is the $m \times m$ identity matrix.  
The initial state is given by $\phi(x_0) = O_b^+ \phi(X)$ in which $O_b^+ = (O_b^TO_b)^{-1}O_b^T$ is the pseudo-inverse of the observability matrix. This pseudo-inverse solution is the least-squares solution that minimizes the distance $d(X)$ in feature space between the collected data $\phi(X)$ and the underlying generating process $O_b \phi(x_0)$ and, such that :

\begin{eqnarray}
d(X)= || \phi(X) - O_b \phi(x_0)||^2 = || \phi(X) - O_b O_b^+\phi(X)||^2 
\label{Eq:dist}
\end{eqnarray}

The problem is that the dynamical system is unknown so that $O_bO_b^+$ is unknown, and the dimension of $\phi(X)$ may be infinite. Thus, direct computation of the distance $d(X)$ is not tractable. 
In the next section, we propose a learning method that allows to estimate $O_bO_b^+$ based on training data and compute (\ref{Eq:dist}) independently of the (potentially infinite) dimensionality of the feature space. 

\section{Dynafit learning}
An estimation of $O_bO_b^+$ in (\ref{Eq:dist}) can be obtained by minimizing the distance in feature space for the training data $X^{(\mu)}  \in \mathbb{R} ^{nN}$, $\mu =1 \cdots p$, which are particular trajectory samples (\ref{Eq:trajectory_vector}) generated by system (\ref{Eq:system1}). 
Let  $X_{train} = (X^{(1)}, X^{(2)}, \cdots, X^{(p)}) \in \mathbb{R} ^{nN \times p}$ be the training set in original space and $\phi(X_{train}) = (\phi(X^{(1)}), \phi(X^{(2)}), \cdots, \phi(X^{(p)})) \in \mathbb{R} ^{l \times p}$ be the training set projected in feature space. The sum of individual distances for the training data in feature space writes as follows : 

\begin{eqnarray}
d(X_{train}) &=& \sum_{\mu=1}^{p} d(X^{(\mu)}) \nonumber \\
&=& \sum_{\mu=1}^{p} || \phi(X^{(\mu)}) - O_b O_b^+\phi(X^{(\mu)})||^2 \nonumber \\
&=& || \underbrace{\phi(X_{train}) - O_b O_b^+ \phi(X_{train})}_\text{E} ||_F^2
\end{eqnarray}

with  
$||E||_F^2=tr(E^T E)$ is the Frobenieus norm and $tr$ is the trace of the matrix.  \\

Let us consider the SVD of $\phi(X_{train})$ 

\begin{eqnarray}
 \phi(X_{train})= U \;\Sigma \;V^T 
 \label{Eq:SVD}
\end{eqnarray}
  
with $U$ and $V$ being the $l \times k$ and $p \times k$ matrices of the left and right singular vectors, respectively.  
The square $k \times k$ diagonal matrix $\Sigma$ has non-zero singular values $\lambda_1, \cdots, \lambda_k$ in the diagonal in decreasing order, i.e. $\lambda_i \geq \lambda_{i+1}$. 

Given that $U^T U = I_k$, 
it is easy to show that $d(X_{train})=0$ if $O_b O_b^+ = U U^T$. Thus, an estimation of $O_b O_b^+$ can be  obtained as $U U^T$ with $U$ being the eigenvectors  
of $\phi(X_{train})\phi(X_{train})^T$ corresponding to non-zero singular values $\lambda_1, \cdots, \lambda_k$. In that case, one has 

\begin{eqnarray}
d(X_{train})  &=& || \phi(X_{train}) - U\; U^T \;\phi(X_{train})||_F^2 \nonumber \\
                   &=& tr \left( \phi(X_{train})^T \phi(X_{train})-\phi(X_{train})^T U U^T \phi(X_{train}) \right)\nonumber\\
                   &&
 \label{Eq:dist_U}
 \end{eqnarray}

as $U U^T$ is an idempotent matrix, i.e. $U U^TU U^T=U U^T$. 
Yet, finding eigenvectors of $\phi(X_{train})\phi(X_{train})^T$, which is $l \times l$,  is not tractable when $l = \infty$. 
To circumvent this problem, we consider a dual optimization.  From (\ref{Eq:SVD}) we have 

\begin{eqnarray}
 U = \phi(X_{train})\;V\;\Sigma^{-1} 
 \label{Eq:Dual}
\end{eqnarray}
  
with $\Sigma^{-1}$ being just the diagonal matrix with $1/\lambda_1, 1/\lambda_2, \cdots, 1/\lambda_k$ on the diagonal. 

Now,  $V$ can be obtained as the eigenvectors corresponding to the non-zero eigenvalues 
of the $p \times p$ Gram matrix $K_{train}=\phi(X_{train})^T \phi(X_{train})$, with $p<<l$ (number of training patterns much smaller than dimensionality of feature space). 
The eigenvalues of $K_{train}$ are the squared singular values $\Sigma^2$ of $\phi(X_{train})$.   
Replacing (\ref{Eq:Dual}) into (\ref{Eq:dist_U}) leads to 

\begin{eqnarray}
d(X_{train}) &=& tr\left( K_{train} - K_{train} H K_{train} \right) 
\label{Eq:dist_training_data} \\
\hbox{with }  H &=& V\;\Sigma^{-2}\;V^T
\label{Eq:H}
\end{eqnarray}

$K_{train}$ is the $p \times p$ kernel gram matrix with entries $k(X^{(i)},X^{(j)}) =\phi(X^{(i)})^T \phi(X^{(j)})$. 
It is worth noting that the kernel function $k(\cdot)$ expresses an inner product in a high-dimensional feature space 
so that the kernel trick from kernel machines can be used to avoid the explicit mapping $\phi$ \citep{Burges_1998, Scholkopf_2000}. \\

Thus, the learning process consists in the eigendecomposition of $K_{train}$ to extract the eigenvectors $V$ and eigenvalues $\Sigma^2$ and compute matrix $H$ (Eq. \ref{Eq:H}). The total distance for the entire training set, i.e. $ \sum_{\mu=1}^{p} d(X^{(\mu)})$, is obtained by Eq. (\ref{Eq:dist_training_data}). 
Individual distances $d(X^{(\mu)})$for each training data $X^{(\mu)}$ are obtained by replacing the trace operator $tr(\cdot)$ in Eq. (\ref{Eq:dist_training_data}) with the function $diag(\cdot)$ that returns the diagonal elements of the matrix. 

\section{Dynafit inference}

Let us consider a set of test data $X_{test} = (X'^{(1)}, X'^{(2)}, \cdots, X'^{(q)}) \in \mathbb{R} ^{nN \times q}$. The total distance on the test set writes 

\begin{eqnarray}
d(X_{test}) = tr\left( K_{test} - K_{test,train} H K^T_{test,train} \right)
\label{Eq:dist_test_data}
\end{eqnarray}

$K_{test}$ is the $q \times q$ matrix with entries $k(X'^{(i)},X'^{(j)}), i,j=1 \cdots q$, and $K_{test,train}$ is the $q \times p$ matrix with entries  $k(X'^{(i)},X^{(j)}), i=1 \cdots q, j=1 \cdots p$. \\

Individual distances $d(X'^{(\mu)})$ for each test sample $X'^{(\mu)}$, $\mu=1 \cdots q$, are obtained by replacing the trace operator $tr(\cdot)$ in Eq. (\ref{Eq:dist_test_data}) with the function $diag(\cdot)$ that returns the diagonal elements of the matrix. 
The class of a test sample $X'^{(\mu)}$ is obtained as follows : 

\begin{itemize}
\item In one-class classification,  the unique class represents the normal operating mode of the system. Unusual events (e.g. fault events) are detected whenever  $d(X'^{(\mu)})$ exceeds a predetermined threshold (deviation from the normal operating mode). \\

\item In multi-class classification, different classes represent different dynamics leading to one distance per class.  The class of $X'^{(\mu)}$ is determined as the one with the smallest distance, that is, class $i$  if $d_i(X'^{(\mu)})<d_j(X'^{(\mu)}), \forall j \neq i$.   
\end{itemize}

\section{Kernel engineering}

Dynafit has been written in terms of kernel functions that correspond to inner products in some high  dimensional (potentially infinite) feature space. 
When the dynamics is unknown, general kernel functions can be used such that the polynomial kernel
 $k(X,Y)=(1+X^TY)^d$ or the Gaussian kernel $k(X,Y)=\exp(-||X-Y||^2/\sigma^2)$. When the dynamics is partially known, specific kernels can be designed to incorporate prior knowledge. \\
 
Here, we illustrate kernel engineering by considering the logistic map, whose state evolves according to the discrete dynamics :  
\begin{eqnarray}
x_{k+1} &=& r x_{k} (1-x_k)\nonumber
\end{eqnarray}
with $x \in [0,1)$. To enforce this domain, we clamp the dynamics to $x=1-\epsilon$  whenever $x=1$, where $\epsilon$ is a very small positive quantity. 
The dynamics is partially known as the bifurcation parameter $r$ is unknown. Depending on the value of $r$, the sequence may settle down to a regular regime (fixed point, periodic orbit) or may produce aperiodic chaos. 
As shown in \citep{Brunton_etal_2016}, the Koopman operator for the logistic map can be represented in a polynomial basis with augmented state :
\begin{eqnarray}
\phi(x) = \begin{pmatrix}
x\\ x^2\\x^3\\ \vdots  
 \end{pmatrix}
\end{eqnarray}

In infinite-dimensional feature space, the state has linear dynamics $\phi( x_{k+1}) = A_\infty \phi( x_k)$ with  : 

\begin{eqnarray}
A_\infty = \begin{pmatrix}
r\hspace{.5cm} & -r\hspace{.5cm}&0\hspace{.5cm}&0\hspace{.5cm}&0\hspace{.5cm}&\cdots\hspace{.5cm}&\cdots\hspace{.5cm}&\cdots\hspace{.5cm}\\ 
0\hspace{.5cm} & r^2\hspace{.5cm}&-2r^2\hspace{.5cm}&r^2\hspace{.5cm}&0\hspace{.5cm}&\cdots\hspace{.5cm}&\cdots\hspace{.5cm}&\cdots\hspace{.5cm}\\
0\hspace{.5cm} & 0\hspace{.5cm}&r^3\hspace{.5cm}&-3r^3\hspace{.5cm}&3r^3&r^3&0&\cdots\hspace{.5cm}\\
\vdots\hspace{.5cm} &\vdots\hspace{.5cm}&\vdots\hspace{.5cm}&\vdots\hspace{.5cm}&\vdots\hspace{.5cm}&\vdots&\vdots&\ddots
\end{pmatrix}
\nonumber
\end{eqnarray}

As noted in \citep{Brunton_etal_2016} the rows of the transition matrix $A_\infty$ are related to the rows of Pascal's triangle. 

Let us consider truncating the state to degree $\omega$ (this constraint will be relaxed later) so that $\phi( x_{k+1}) \approx A_\omega \phi( x_k)$ with 

\begin{eqnarray}
\phi(x) = \begin{pmatrix}
x\\ x^2\\ \vdots\\x^\omega
 \end{pmatrix}
\label{Eq:polynomial}
\end{eqnarray}

The
truncated system, with $\omega \times \omega$ transition matrix $A_\omega$,  approximates the true dynamics only for very large $\omega$. 
Yet, for large $\omega$, the feature space becomes intractably large which motivates the use of kernels. \\

Let us consider trajectory vectors $ \phi(X)$ made of $N$ consecutive observations (Eq \ref{Eq:observation_vector}). 
The kernel function expresses an inner product in feature space between two trajectories; that is : 

\begin{eqnarray}
 k(X, Y) = \phi(X)^T \phi(Y) 
\label{Eq:K_XY}
\end{eqnarray}

Using (\ref{Eq:polynomial}), one has 

\begin{eqnarray}
 k(X, Y) = \sum_{k=0}^{N-1}  \left(\sum_{\xi=1}^\omega (x_k y_k)^\xi \right)  \nonumber
\end{eqnarray}

By identifying a geometric series for the summation between brackets, the kernel function simplifies to  :   

\begin{eqnarray}
k(X, Y) = \sum_{k=0}^{N-1} \frac{ x_k y_k(1- (x_k y_k)^\omega)}{1- x_k y_k} \nonumber
\end{eqnarray}

For $\omega \rightarrow \infty$, one has $ (x_k y_k)^\omega \rightarrow 0$ as $x_k, y_k \in [0,1)$. Thus, the logistic map kernel writes as follows 

 \begin{eqnarray}
k(X, Y) = \sum_{k=0}^{N-1} \frac{ x_k y_k}{1- x_k y_k}
\label{Eq: logistic_kernel}
\end{eqnarray}

Evaluating the logistic map kernel (\ref{Eq: logistic_kernel}) is much cheaper than calculating the mapping $\phi$ using (\ref{Eq:polynomial}) and the inner product (\ref{Eq:K_XY}) in feature space of high dimensionality. 
Also, the logistic map kernel (\ref{Eq: logistic_kernel}) exactly represents the infinite-dimensional dynamics as $\omega=\infty$ and not the approximated dynamics of a truncated system. 

\section{Results}
The Matlab code to reproduce the experiments can be found in the following Github repository : \url{https://github.com/dynafit-sketch/dynafit}.  

\subsection*{Chaos detection}

In the past, chaos detection has been applied to neural data, meteorological data, financial and stock market, among others. 
Detecting chaotic regimes is critical for understanding and modeling a dynamical system, for predicting its future behavior and performing decision-making. 
With chaotic dynamics, one will never be able to predict the data beyond a brief interval determined by the  Lyapunov exponent (example 4-5 days for weather prediction).  
Here, we take the logistic map as an example of chaotic system and we use Dynafit with the logistic map kernel (\ref{Eq: logistic_kernel}) to discriminate between regular and chaotic regimes. To compare with deep learning  we used the same dataset and parameter settings as in \citep{Barrio_etal_2023}. 
In this dataset\footnote{ \url{https://doi.org/10.17632/k4x675k5dm.2}}, trajectories of $N=1000$ samples are labeled as regular or chaotic depending of their Lyapunov exponent.  
We trained two metrics, $d_1$  for regular and $d_2$ for chaotic, using 2000 training data for each class. 
After training, Dynafit classified the test data as regular when $d_1<d_2$ and chaotic otherwise.
 Table 1 sumarizes the results obtained as compared to those reported in \citep{Barrio_etal_2023} for Multi-Layer Perceptron (MLP), Cellular Neural Network (CNN) and Long Short-Term Memory (LSTM).  
 We note that Dynafit outperforms most of the deep learning classifiers. Training accuracy for Dynafit is higher than for  MLP, CNN and LSTM. Test accuracy for Dynafit is better than for MLP and LSTM and similar to that of CNN (99.4\%). 
 In addition, Training is very fast for Dynafit, that is less than 10 sec on a standard laptop DELL latitude 7410 for computing the $2000\times2000$ kernel matrices and solving the eigenvalue problems for the two classes (regular and chaotic).   

\begin{table}[h!]
  \begin{center}
    \label{tab:tableX}
    \begin{tabular}{c|c|c}
      &Train acc. (in \%) & Test acc. (in \%)\\
      \hline
 Dynafit & 100 & 99.4 \\
      \hline
      MLP & 99.302±0.161 & 97.865±0.098\\
      CNN &  99.383±0.335 & 99.410±0.239\\
      LSTM & 97.979±0.584 & 98.565±0.279\\
      \hline
   \end{tabular}
    \vspace{.5cm} 
\caption{Logistic map experiments. Train and test accuracy on the training set (2000 samples) and the test set (1000 samples) for Dynafit as compared to deep learning. % The best classifier in each condition is indicated in bold.  
Dynafit is trained with the logistic map kernel given in Eq. (\ref{Eq: logistic_kernel}). 
Results for  MLP, CNN and LSTM are taken from \citep{Barrio_etal_2023}.   }
  \end{center}
\end{table}

\subsection*{Character recognition from pen trajectory}

We consider a real dataset from the UCI Machine learning archive\footnote{ 
\url{https://archive.ics.uci.edu/dataset/175/character+trajectories}}. The dataset consists of 2858 handwriting characters of 20 classes (`a,b,c,d,e,g,h,l,m,n,o,p,q,r,s,u,v,w,y,z') collected using a WACOM pen tablet. The sensor data consists in pen trajectories with three components recorded at 200 Hz: pen velocity in x-position (mm/s), pen velocity in y-position (mm/s) and rate of the pen tip force (N/s). Figure 2 shows examples of characters reconstructed using the first $N$=10, 20, 50 and 100 pen trajectory data. We used Dynafit with one distance metric per class. The training set consisted of 10\% of the recordings, randomly sampled from the dataset, and the test set consisted of the remaining 90\%. Training and test were repeated 100 times. 
Table 2 sumarizes the results obtained with Dynafit as compared to SVM. 
Standard SVM classification is performed by following the one-vs-all strategy that splits the multiclass problem into one binary classification problem per class. 
We note that Dynafit outperforms SVM both in terms of test accuracy and training time with significant speedup. 
In SVM, we solve one quadratic programing problem per class that involves all the data. 
In Dynafit, we solve one eigenvalue problem per class that involves merely the data from one class. This simpler problem makes Dynafit up to 260 times faster than the standard SVM.  
 
\begin{figure}[htbp]
\centerline{\includegraphics[width=10cm]{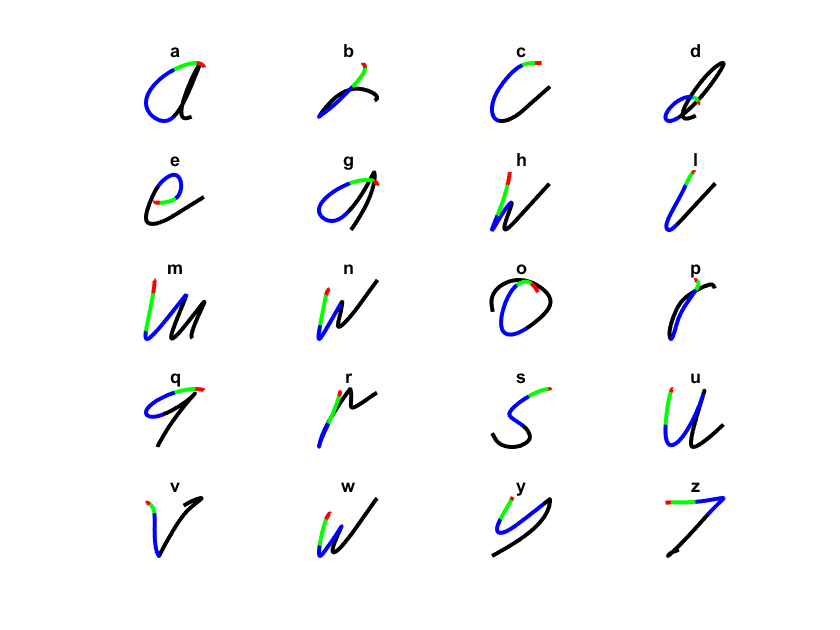}}
\caption{Reproduction of characters with the first $N=$10 (plots in red), 20 (plots in green), 50 (plots in blue) and 100 (plots in black) sensor data sampled during the pen trajectory. The data are the pen velocity in x-position (mm/s), pen velocity in y-position (mm/s) and rate of the pen tip force (N/s) recorded at 200 Hz.}
\end{figure}

\begin{table}[htbp]
  \begin{center}
    \label{tab:table1}
    \begin{tabular}{c|c|c c|c c|}
 &  & {\bf Dynafit}  &  & {\bf SVM} &  \\   
 Kernel   & N & Test Acc. & Train time  &  Test Acc. & Train time \\
 &  &  (in \%) & (in ms) &  (in \%) & (in ms) \\
 %      \textbf{Kernel function} & \textbf{N} & \textbf{Acc. (in \%)} \textbf{Train (in sec)} &  \textbf{Acc. (in \%)} \textbf{Train (in sec)} \\
       \hline
      \multirow{4}{*}{polynomial} & 10 & 37.6$\pm$1.1& 1.1$\pm$0.5&35.3 $\pm$1.2&268.2$\pm$41.8\\ 
     & 20 &50.9$\pm$1.1&1.2$\pm$0.5&51.5$\pm$1.4&319.9$\pm$57.3\\ 
      & 50 & 81.3$\pm$1.0&1.6$\pm$0.7&80.0$\pm$1.3&188.1$\pm$22.3\\ 
      & 100 & 94.2$\pm$0.7&1.8$\pm$0.8&93.9$\pm$0.8&174.3$\pm$17.5\\ 
      \hline     
      \multirow{4}{*}{gaussian} & 10 &38.5$\pm$1.0& 4.1$\pm$1.3&31.5 $\pm$1.3&147.2$\pm$18.3\\ 
     & 20 &52.5$\pm$1.1&5.0$\pm$1.2&49.3$\pm$1.4&156.3$\pm$18.9\\ 
     & 50 & 82.5$\pm$1.0&7.5$\pm$1.6&78.3$\pm$1.3&159.8$\pm$14.8\\ 
    & 100 & 93.7$\pm$0.7&8.9$\pm$1.3&91.3$\pm$0.9&175.5$\pm$16.4\ 
      
          \end{tabular}
  \end{center}
  \caption{Test accuracy on the test set with the first $N=$10, 20, 50 or 100 sampled points in the pen trajectory. The polynomial kernel has degree $d=2$ for all $N$. The 
  Gaussian kernel has width $\sigma=$ 2.9, 4.2, 7.5 and 15 for $N=$10, 20, 50 and 100, respectively. 
  Results are represented as means$\pm$S.D estimated over 100 trials.  `Train time' is the time in millisecond for training the classifier in each trial.  }
\end{table}

\subsection*{Dynamic texture classification}
Dynamic textures (also called temporal textures) are time-varying visual patterns that are modeled with linear state-space representations \citep{Doretto_etal_2003}. Here, we use Dynafit to  classify dynamic textures. The UCLA dynamic texture dataset \citep{Saisan_etal_2001} consists of 50 classes of different textures, each class having 4 sequences of 75 frames of 48 x 48 pixels. 
For each class, we used three sequences for training and the remaining one for testing. 
Fig. 3 shows the first images from the test sequence for each class. 
Using Dynafit, we obtain a test accuracy of 78\% and 86\% with a polynomial kernel of degree 1 and 3, respectively. These performance are well above the chance level of 2\% and are comparable to those obtained in \citep{Saisan_etal_2001} with the Martin distance \citep{Martin_2000,DeCock_DeMoor_2000}, i.e. 71\% and 89.5\% with features extracted from  independant component analysis and principal component analysis, respectively. 

\begin{figure}[htbp]
\centerline{
\includegraphics[width=8cm]{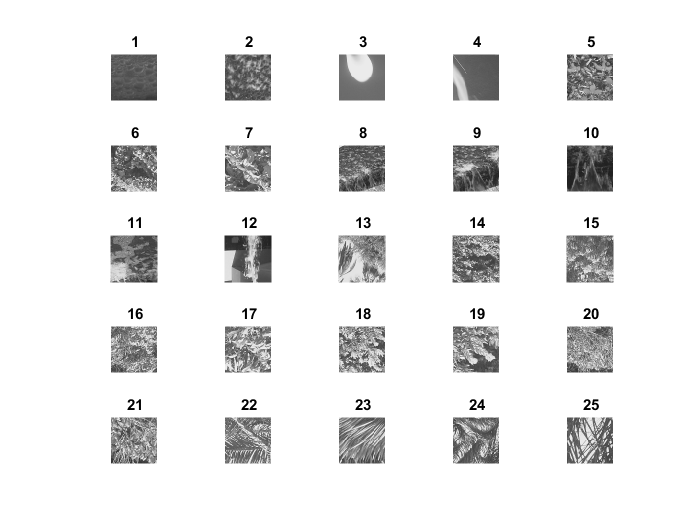}
\hspace{-1cm}
\includegraphics[width=8cm]{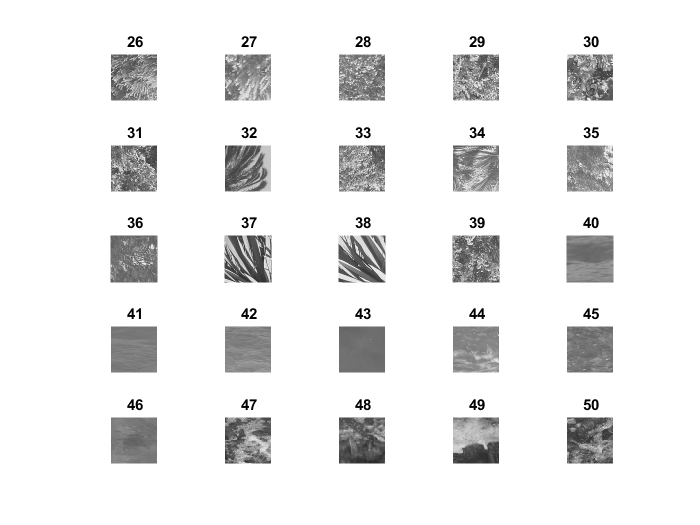}
}
\caption{Image samples from the UCLA dynamic texture dataset (50 classes of different dynamic textures). Each pannel corresponds to the first image in the test sequence of a given class.}
\end{figure}

\section{Conclusion}

We introduced a data-driven distance metric for the classification of trajectories and sequences generated by nonlinear dynamical systems. The metric is based on a kernel-induced projection that approximates the Koopman operator, thereby enabling a global linearization of the dynamics in a potentially infinite-dimensional feature space.
As a proof of concept, we analytically derived a kernel function from the polynomial Koopman basis that linearizes the dynamics of the logistic map. 
Using this kernel, the proposed metric successfully discriminates between regular and chaotic regimes, achieving performance comparable to—and in some cases exceeding—that of deep learning approaches.
Unlike deep learning methods, the proposed framework does not require the end-to-end training of large parametric models. Instead, it relies on solving single-class eigenvalue problems, resulting in a computationally efficient learning procedure. This advantage was demonstrated on a character trajectory benchmark, where the proposed approach achieved a 260× speedup compared with SVM-based classification.
It should be emphasized, however, that the objective of the experimental evaluation was not to provide an exhaustive comparison with state-of-the-art machine learning or deep learning techniques. Rather, the goal was to demonstrate that the proposed Koopman-based metric captures discriminative information about the underlying dynamics and can therefore serve as an effective tool for trajectory classification.
Future work will focus on further reducing the computational requirements of the method to facilitate deployment in resource-constrained edge-learning environments. In particular, the Nyström approximation \cite{Williams_Seeger_2000} will be investigated as a means of efficiently scaling the approach to large datasets and high-dimensional problems.

\acks{This work was funded by France 2030 under the investment program ANR-20-PCPA-0007 Pherosensor and by ANR-23-CE51-0037 Bucolyc.}

\bibliography{Martinez_arXiv_2026}

@article{Williams_etal_2015a,
title = {A Data–Driven Approximation of the Koopman Operator: Extending Dynamic Mode Decomposition},
journal = { J Nonlinear Sci},
volume = {25},
pages = {1307–1346}, 
year = {2015},
doi = {10.1007/s00332-015-9258-5},
author = {M.O.  Williams and I.G.  Kevrekidis and C.W. Rowley}
}

@article{Williams_etal_2015b,
title = {A kernel-based method for data-driven koopman spectral analysis},
journal = {Journal of Computational Dynamics},
volume = {2},
number = {2},
pages = {247-265}, 
year = {2015},
issn = {2158-2491},
doi = {10.3934/jcd.2015005},
url = {https://www.aimsciences.org/article/id/ce535396-f8fe-4aa1-b6de-baaf986f6193},
author = {M.O.  Williams and C.W. Rowley and I.G.  Kevrekidis},
}

@article{Mezic_Banaszuk_2004,
title = {Comparison of systems with complex behavior},
journal = {Physica D: Nonlinear Phenomena},
volume = {197},
number = {1},
pages = {101-133},
year = {2004},
issn = {0167-2789},
doi = {https://doi.org/10.1016/j.physd.2004.06.015},
url = {https://www.sciencedirect.com/science/article/pii/S0167278904002507},
author = {I. Mezić and A. Banaszuk}
}

@article{Mezic_2016,
title = {On Comparison of Dynamics of Dissipative and Finite-Time Systems Using Koopman Operator Methods},
journal = {IFAC-PapersOnLine},
volume = {49},
number = {18},
pages = {454-461},
year = {2016},
note = {10th IFAC Symposium on Nonlinear Control Systems NOLCOS 2016},
issn = {2405-8963},
doi = {https://doi.org/10.1016/j.ifacol.2016.10.207},
url = {https://www.sciencedirect.com/science/article/pii/S2405896316317876},
author = {I. Mezić},
}

@article{Mezic_2005,
title = {Spectral Properties of Dynamical Systems, Model Reduction and Decompositions},
journal = {Nonlinear Dynamics},
volume = {41},
number = {1},
pages = {309-325},
year = {2005},
author = {I. Mezić}
}

@article{Mezic_2013,
   author = {I. Mezić},
   title = "Analysis of Fluid Flows via Spectral Properties of the Koopman Operator", 
   journal= "Annual Review of Fluid Mechanics",
   year = "2013",
   volume = "45",
   number = "Volume 45, 2013",
   pages = "357-378",
   doi = "https://doi.org/10.1146/annurev-fluid-011212-140652",
   url = "https://www.annualreviews.org/content/journals/10.1146/annurev-fluid-011212-140652"
  }

@article{Ravindran_2025,
  title={Cutting AI down to size},
  author={S. Ravindran},
  journal={Science},
  volume={387},
  number={6736},
  pages={818--821},
  year={2025}
}

@article{Ray_2022,
  title={A review on TinyML: State-of-the-art and prospects},
  author={P.P. Ray},
  journal={Journal of King Saud University - Computer and Information Sciences},
  volume={34},
  number={4},
  pages={1595--1623},
  year={2022}
}

@article{LeCun_etal_2015,
  title={Deep Learning},
  author={Y. LeCun and Y. Bengio and G. Hinton},
  journal={Nature},
  volume={521},
  pages={436--444},
  year={2015}
}

@article{Cover_Hart_1967,
  title={Nearest neighbor pattern classification},
  author={T. Cover and P. Hart},
  journal={IEEE Transactions on Information Theory},
  volume={13},
  number={1},
  pages={21--27},
  year={1967}
}

@article{Lloyd_1982,
  title={Least squares quantization in PCM},
  author={S. Lloyd},
  journal={IEEE Transactions on Information Theory},
  volume={28},
  number={2},
  pages={129--137},
  year={1982}
}

@article{Parzen_1962, 
  title={ On  estimation  of a  probability  density  function  and  mode},
  author={E. Parzen},
  journal={IAnn.  Math.   Stat},
  volume={33},
  pages={1065--1076},
  year={1962}
}

@article{Abanda_etal_2019, 
  title={ A review on distance based time series classification},
  author={A. Abanda and U. Mori and J.A. Lozano},
  journal={Data Min. Knowl. Disc.},
  volume={33},
  pages={378--412},
  year={2019}
}

@article{Burges_1998, 
  title={ A Tutorial on Support Vector Machines for Pattern Recognition},
  author={C.J. Burges},
  journal={Data Mining and Knowledge Discovery},
  volume={2},
  pages={121--167},
  year={1998}
}

@article{Scholkopf_2000, 
  title={The kernel trick for distances},
  author={B. Scholkopf},
  journal={Advances in Neural Information Processing Systems (NIPS)},
  volume={2},
  year={2000},
  publisher={MIT Press}
}

@article{Battistelli_Tesi_2021,
  title={Classification for Dynamical Systems: Model-Based and Data-Driven Approaches},
  author={G. Battistelli and P. Tesi},
  journal={IEEE Transactions on automatic Control},
  volume={6},
  pages={1741--1748},
  year={2021},
  publisher={IEEE}
}

@article{Martinez_Boutayeb_2023,
  title={Classification for Dynamical Systems: Model-Based and Data-Driven Approaches},
  author={D. Martinez and M. Boutayeb},
  journal={IEEE Sensors Journal},
  volume={23},
  number={22},
  pages={28454--28461},
  year={2023},
  publisher={IEEE}
}

@article{Patton_Chen_1991,
  title={A review of parity space approaches to fault diagnosis},
  author={R.J. Patton and J. Chen},
  journal={IFAC Proceedings},
  volume={24},
  number={6},
  pages={65--81},
  year={1991}
}

@article{Martinez_Boutayeb_2024,
  title={Nullspace-based metric for classification of dynamical systems and sensors},
  author={D. Martinez and M. Boutayeb},
  journal={Lecture Notes in Computer Science, Springer, Cham, },
  volume={15016},
  pages={18--29},
  year={2024}
}

@article{Koopman_1931,
  title={Hamiltonian Systems and Transformation in Hilbert Space},
  author={B. Koopman},
  journal={Proceedings of the National Academy of Sciences},
  volume={17},
  number={5},
  pages={315--318},
  year={1931}
}

@article{Koopman_Neumann_1932,
  title={Hamiltonian Systems and Transformation in Hilbert Space},
  author={B. Koopman and J. Neumann},
  journal={Proceedings of the National Academy of Sciences},
  volume={18},
  number={3},
  pages={255},
  year={1932}
}

@article{Brunton_etal_2016,
  title={Koopman Invariant Subspaces and Finite Linear Representations of Nonlinear Dynamical Systems for Control},
  author={J.L. Brunton and B.W. Brunton and J.L. Proctor and J.N. Kutz},
  journal={PLoS One},
  year={2016}
}

@article{Barrio_etal_2023,
  title={Deep Learning for Chaos Detection},
 author={R. Barrio and A. Lozano Rojo and A. Mayora-Cebollero and C. Mayora-Cebollero and A. Miguel and A. Ortega and S. Serrano and R. Vigara},
    journal={Chaos},
  volume={33},
  number={073146},
  year={2023}
}

@article{Williams_Seeger_2000, 
  title={Using the Nystr\"{o}m Method to Speed Up Kernel Machines},
  author={C. Williams and M. Seeger},
  journal={Advances in Neural Information Processing Systems (NIPS)},
  volume={13},
  year={2000},
  publisher={MIT Press}
}

@article{DeCock_DeMoor_2000, 
  title={Subspace angles and distances between ARMA models},
  author={K. De Cock and B. De Moor},
  journal={Systems \& Control Letters},
  volume={46},
  number={4},
  pages={265--270},
  year={2002}
}

@article{Martin_2000, 
  title={A metric for ARMA processes},
  author={R.J. Martin},
  journal={IEEE Transactions on Signal Processing},
  volume={48},
  number={4},
  pages={1164--1170},
  year={2000}
}

@article{Doretto_etal_2003, 
  title={Dynamic textures},
  author={G. Doretto and A. Chiuso and Y.N. Wu and S. Soatto},
  journal={Intl. J. Computer Vision},
  volume={51},
  number={2},
  pages={91--109},
  year={2003}
}

@article{Saisan_etal_2001, 
  title={Dynamic texture recognition},
  author={P. Saisan and G. Doretto and Y.N. Wu and S. Soatto},
  journal={in IEEE Conf. CVPR},
  volume={2},
  pages={58--63},
  year={2001}
}

@article{Wolf_Shashua_2003, 
  title={Learning over sets using kernel principal angles},
  author={L. Wolf and A. Shashua},
  journal={Journal of Machine Learning Research},
  volume={4},
  pages={913--931},
  year={2003}
}

@article{Vishwanathan_etal_2007, 
  title={Binet-Cauchy kernels on dynamical systems and its application to the analysis of dynamic scenes},
  author={S.V.N. Vishwanathan and A.J. Smola and R. Vidal},
  journal={Int. Journal of Computer Vision},
  volume={73},
  pages={95--119},
  year={2007}
}

\end{document}